\documentclass[letterpaper, 10 pt, conference]{ieeeconf}

\IEEEoverridecommandlockouts

\usepackage{cite}
\usepackage{amsmath,amssymb,amsfonts}
\usepackage{algorithm,algpseudocode}
\usepackage{graphicx}
\usepackage{textcomp}
\usepackage{url}
\usepackage{hyperref}
\usepackage{multirow}

\def\showtodo{0}

\usepackage{todonotes}
\usepackage{tabularx}
\usepackage{xcolor}

\newtheorem{lemma}{Lemma}
\newtheorem{definition}{Definition}

\newtheorem{property}{Property}

\newtheorem{assumption}{Assumption}

\newtheorem{problem}{Problem}
\newtheorem{remark}{\textbf{Remark}}

\newcommand{\rr}{\mathop{{\rm I}\mskip-4.0mu{\rm R}}\nolimits}

\newcommand{\Z}{\mathop{{\rm Z}\mskip-7.0mu{\rm Z}}\nolimits}

\def\BibTeX{{\rm B\kern-.05em{\sc i\kern-.025em b}\kern-.08em
		T\kern-.1667em\lower.7ex\hbox{E}\kern-.125emX}}

\DeclareMathOperator*{\argmin}{arg\,min}

\DeclareRobustCommand{\IEEEauthorrefmark}[1]{\smash{\textsuperscript{\footnotesize #1}}}

\allowdisplaybreaks

\title{\LARGE \bf
Collision-Free Platooning of Mobile Robots through a Set-Theoretic  Predictive Control Approach}

\author{Suryaprakash Rajkumar\IEEEauthorrefmark{1}, Cristian Tiriolo\IEEEauthorrefmark{1} and Walter Lucia
\thanks{This work was supported by the Natural Sciences and Engineering Research Council of Canada (NSERC).}
\thanks{Suryaprakash Rajkumar, Cristian Tiriolo and Walter Lucia are with the Concordia Institute for Information Systems Engineering (CIISE), Concordia University, Montreal, QC, H3G 1M8, CANADA, {\tt\small \{suryaprakash.rajkumar, cristian.tiriolo, walter.lucia\}@concordia.ca}%, {\tt\small cristian.tiriolo@concordia.ca}, {\tt\small walter.lucia@concordia.ca}}
}
\thanks{ \IEEEauthorrefmark{1}These authors contributed equally to this work.}
}

\begin{document}

\maketitle
\thispagestyle{empty}
\pagestyle{empty}

%%%%%%%%%%%%%%%%%%%%%%%%%%%%%%%%%%%%%%%%%%%%%%%%%%%%%%%%%%%%%%%%%%%%%%%%%%%%%%%%
\begin{abstract}
This paper proposes a control solution to achieve collision-free platooning control of input-constrained mobile robots. The platooning policy is based on a leader-follower approach where the leader tracks a reference trajectory while followers track the leader's pose with an inter-agent delay. First, the leader and the follower kinematic models are feedback linearized and the platoon's error dynamics and input constraints characterized.
Then, a set-theoretic model predictive control strategy is proposed to address the platooning trajectory tracking control problem.  An ad-hoc collision avoidance policy is also proposed to guarantee collision avoidance amongst the agents. Finally, the effectiveness of the proposed control architecture is validated through experiments performed on a formation of Khepera IV differential drive robots. 
\end{abstract}
 
% \begin{IEEEkeywords}
%     Platooning control, wheeled bobile robots, set-theoretic model predictive control, robust control.
% \end{IEEEkeywords}

\section{Introduction}
In the quest for autonomous vehicles,  autonomous vehicular platooning is a practical option that offers various advantages, including increased safety, reduced drag, and greater performance in comparison to a single autonomous vehicle.
Platoon in vehicular technologies is often defined as a group of vehicles traversing in a coordinated manner while communicating with each other and by using autonomous driving technology \cite{platoon_book}. 

In the literature, different solutions have been proposed to control platoons of mobile robots (see \cite{Christian_review_platoon} and references therein). 
%
% In \cite{ACC_wang},  the problem is solved in a centralized fashion via nonlinear controllers, while in \cite{RAL_Kostas} a collision-free solution is developed; on the other hand, the authors of \cite{KLANCAR2011} propose a decentralized constant-time approach where an inter-vehicle delay policy is designed to track a given reference trajectory while maintaining a platoon structure. 
% Although appealing, the above-mentioned
Most of the control solutions available in the literature share as a common drawback the incapability to guarantee input constraint fulfillment, i.e., physical limitations on the control signals of the vehicles are not directly considered in the control design. Unfortunately, failing to address input constraints may lead to undesired saturation phenomena, loss of tracking performance, and in the worst case collisions between agents. 
 Model Predictive Control (MPC) has been successfully applied to the control of vehicular platoons thanks to its peculiar capability to incorporate stability requirements, tracking performance, and state and input constraints directly in the control design. 

 Nonlinear MPC has been investigated to solve platooning control problems for autonomous vehicles, (see \cite{liu2022cooperative} and references therein), however, its high computational burdens represent a major obstacle for real-time implementations. Furthermore, nonlinear formulations are in general nonconvex and they suffer from local minima problems \cite{Subramani_mpc}. 
Conversely, simpler linear MPC formulations are more computationally affordable, and preferred for real-time applications. 
In \cite{Franze_multi}, the linear dual-mode Set-Theoretic MPC (STMPC), first developed in \cite{angeli2008ellipsoidal} to stabilize a general linear system subject to bounded disturbance, has been applied to the considered platooning control problem for mobile robots. 
Such a control architecture combines the concept of control invariance and N-step controllability, to formulate a convex and recursively feasible optimization. 
However, the strategy in \cite{Franze_multi} deals with agents described by linear time-invariant models. 
The latter may constitute a restrictive assumption especially when time-varying reference trajectories are considered.

Feedback Linearization (FL) is a well-established linearization technique capable of transforming a nonlinear time-invariant model into an equivalent linear one \cite{oriolo2002wmr}.  However, simplifying the robot's dynamics comes at the cost of increasing the complexity of its input constraints. As a matter of fact, when FL is applied to linearize mobile robot's kinematics, even simple-box-like input constraint sets transform into time-varying orientation-dependent sets \cite{Tiriolo_TCST}. Consequently, if FL is exploited for model predictions, the resulting MPC formulation is inevitably nonlinear and nonconvex. 
In \cite{tiriolo2023set}, a worst-case approximation of the time-varying input constraint set (first formulated in \cite{Tiriolo_TCST}) is exploited to solve the trajectory tracking problem for a single mobile robot.  Specifically, offline, a family of Robust One-Step Controllable (ROSC) sets is built by considering such a worst-case approximation of the input constraint set. 
Then, online, the optimization problem is relaxed by exploiting the knowledge of the actual time-varying polyhedral constraint. The strength of such an approach is the formulation of recursively feasible quadratic optimizations, which ensure input constraint fulfillment and bounded tracking error.

\if\showtodo 1
\todo[inline,color=green]{The sentence below might be questionable. ST-MPC \cite{ANGELI20083113} is for linear system and you claim that is better than non linear MPC. Do you actually mean for application to feeback linearized vehicles? If yes, mention our paper  before to support the claim, specifically for vehicles - Corrected}
\todo[inline]{My comment above is not solved. We cannot compare STMPC with nonlinear MPC. This makes sense only if e.g., in the sense sentence we clarify that we refer to nonlinear MPC for vehicles vs STMPC for the feedbacklinearized vehicle.}
\fi

The contribution of the proposed leader-follower platoon strategy can be summarized as follows. It characterizes the error dynamics in a leader-follower formation, modeling them as linear systems subject to bounded disturbances. 
% Worst-case scenarios are considered for input constraints and disturbances on each agent.
A customized Set-Theoretic Model Predictive Control (STMPC) is designed to ensure constraint fulfillment and finite-time bounded error for leader and follower vehicles. The strategy proposes a collision avoidance policy based on set-theoretic reachability arguments to avoid collisions through adaptive inter-vehicle delays. Experiments with three Khepera IV robots validate the effectiveness of the proposed approach in a practical setting.
%The proposed experimental comparison is meant to show that failing to directly address input constraints, leads to a loss of tracking performance, for the considered platoon of mobile robots. 

% %\section{Structure of the paper}
% The paper is organized as follows. In section \ref{Prelims_PF}, first some preliminary differential-drive robots modeling concepts and definitions are introduced.  Then the platooning problem of interest is formulated. In section \ref{sec:propo-solution}, first, the feedback linearized error dynamics for the platooning problem are characterized along with the admissible input constraint set. Then, the proposed ST-MPC-based tracking controller is designed. Finally, in Section \ref{sec:Exp_results}, the proposed solution is validated by means of laboratory experiments conducted on a formation of differential-drive robots and its effectiveness contrasted with an alternative strategy in the literature.
% \section{Preliminaries and Problem Formulation}
% \label{Prelims_PF}

\subsection{Preliminaries} 
\begin{definition}\label{def:mink-sum} 
Given two sets $\mathcal{A}$, $\mathcal{B}\subset\rr^n$, their Minkowski sum ($\oplus$) and difference ($\ominus$):
\begin{eqnarray}
    \mathcal{A} \oplus \mathcal{B}&:=&\{a+b: \, a\in\mathcal{A}, \, b\in\mathcal{B} \} \nonumber\\
    \mathcal{A} \ominus \mathcal{B}&:=&\{a \in \rr^n: \, a+b\in \mathcal{A},\,  \forall b\in\mathcal{B} \nonumber \}.
\end{eqnarray}
\end{definition}
\begin{definition}
    Ellipsoidal set $\mathcal{E}(Q)$, with shaping matrix $Q$ and centre at orgin is defined as
        $$
        \mathcal{E}(Q)=\{z \in \rr^n \mid z^TQ^{-1}z \leq 1\}, %Q = Q^{T} > 0,Q\in\rr^{n\times n}
        $$
    where $Q$ is the positive definite for all $\rr^n$.   
\end{definition}

% \begin{property}
%     Given two ellipsoidal sets $\mathcal{E}_{1}$ and $\mathcal{E}_{2}$ with centre at the orgin in the form
% \begin{equation*}
% \begin{array}{rcl}
%      \mathcal{E}_1=\{z \in \rr^2 \mid z^TQ_{1}^{-1}z \leq 1\}, Q_{1} = r_{c1}^2 I\\  
%     \mathcal{E}_2=\{z \in \rr^2 \mid z^TQ_{2}^{-1}z \leq 1\}, Q_{2} = r_{c2}^2 I
% \end{array}
% \end{equation*}
% where $ r_{c1}^2$ and $ r_{c2}^2$ are the radii of $\mathcal{E}_{1}$ and $\mathcal{E}_{2}$ respectively, then the Minkowski sum and difference is defined as,
% \begin{equation*}
% \begin{array}{rcl}
%      \mathcal{E}_1 \oplus \mathcal{E}_2=\{z \in \rr^2 \mid z^TQ_{s}^{-1}z \leq 1\}, Q_{s} = r_{s}^2 I\\  
%     \mathcal{E}_1 \ominus \mathcal{E}_2=\{z \in \rr^2 \mid z^TQ_{d}^{-1}z \leq 1\}, Q_{d} = r_{d}^2 I\\ 
% \end{array}
% \end{equation*}
% where $r_{s} =  r_{c1} + r_{c2}$ while assuming for Minkowski difference, $r_{c1} > r_{c2}$, $r_{d} =  r_{c1} - r_{c2}$.  
% \end{property}
Consider the following discrete linear system:
\begin{equation}\label{eq:gen-lin-sys}
	z(k+1)\!=\!Az(k)+Bu(k)+d(k),\,\,\,	u(k)\in \mathcal{U},\,d(k)\in \mathcal{D}\\
\end{equation}
where $k\in \Z:=\{0,1,\ldots\},$  $z\in\rr^n$, $u\in\rr^m$, $d\in\rr^n$ and $\mathcal{U}\subset\rr^m$, $\mathcal{D}\subset\rr^{n}$ are compact and convex sets containing the origin. 
\begin{definition}
Considering the linear system in \eqref{eq:gen-lin-sys}, Robust one Step Controllable Set (ROSC) from a set $\mathcal{T}^i\subset\rr^n$ is defined as \cite{borrelli_bemporad_morari_2017}:
\begin{equation*}
    \mathcal{T}^{i+1} = \{ z\in\rr^n : z \in ((\mathcal{T}^i \ominus\mathcal{D}) \oplus (-B \cdot \mathcal{U})) \cdot A) \}
\end{equation*}
\end{definition}
\begin{definition}\label{def:RCI}
\label{Def:ROSR}
    Considering the linear system in \eqref{eq:gen-lin-sys}, the Robust One Step Reachable (ROSR) set  from a set $\mathcal{R}_{i}\subset\rr^n$ is defined as \cite{borrelli_bemporad_morari_2017},
    \begin{equation*}
        \mathcal{R}_{i+i} = \{ z\in\rr^n : z \in (( A \cdot \mathcal{R}_i) \oplus (B\cdot \mathcal{U}) \oplus \mathcal{D}\}
    \end{equation*}
\end{definition}
\begin{definition}
    A set is $\mathcal{C}$ is said to be a Robust Control Invariant (RCI) set for the system  \eqref{eq:gen-lin-sys},
    \begin{equation*}
        \forall z\in\mathcal{C}, \, \exists u\in \mathcal{U}:\, Az+Bu+d \in \mathcal{C}, \,\forall d\in\mathcal{D}
    \end{equation*}
\end{definition}
\begin{definition}
    The distance between a set $\mathcal{S}\subset\rr^n$ and a point $p\in\rr^n$ is defined as:
    $$
    dist(p,\mathcal{S})=\inf_{s\in\mathcal{S}}\|p-s\|_2
    $$
\end{definition}
\subsection{Robot Modelling}
% \todo[inline,color=yellow]{Figure is updated with D and $x^i, y^i$. Please verify}
% \begin{figure}[!h]
% 	\centering
% \includegraphics[width=0.7\linewidth]{Figure/vehicle-models}
% 	\caption{\textit{(a)} differential-drive, \textit{(b)} unicycle.}
% 	\label{fig:vehicles-model}
% \end{figure}
%
Let's consider a differential-drive robot described by the following discrete-time nonlinear kinematic model :
\begin{equation}\label{eq:diffdrive-discrete}
    \begin{array}{rcl}
        x^i(k+1)&=&x^i(k)+T_s \frac{R}{2}\left(\omega_R^i(k)+\omega_L^i(k)\right)\cos\theta^i(k)\\
        y^i(k+1)&=&y^i(k)+T_s \frac{R}{2}\left(\omega_R^i(k)+\omega_L^i(k)\right)\sin \theta^i(k)\\
        \theta^i(k+1)&=&\theta^i(k)+T_s\frac{R}{D}(\omega_R^i(k)-\omega_L^i(k))
    \end{array}
\end{equation}
Where $T_s>0$ is the sampling time, $q^i=[x^i,y^i,\theta^i]^T$ 
is the pose of the geometric center of the robot. $R$, and $D$ are the wheel radius and axis length of the robot, respectively. The left and the right wheel angular speeds $\omega^i_R,\omega^i_L \in \rr$ are the control inputs of the system.
%is the robot's pose, i.e.,  the position of the centre of mass of the robot, and its orientation,  $R$ is the whee l radius and $D$ is the axis length. On the other hand, $\omega^R_i,\omega^L_i \in \rr$ are the control inputs, i.e., the right and left wheels' angular velocities, respectively.
Furthermore, the control inputs are subject to box-like constraints, i.e., the set of admissible wheels' angular speed for the differential drive:
\begin{equation}\label{eq:diff-drive-constraints}
\begin{array}{c}
   \mathcal{U}_d=\{[\omega_R^i,\omega_L^i]^T\in\rr^2:\, H_d\left[\omega_R^i,\omega_L^i \right]^T\leq \bf{1}\},  \\
     H_d=\left[\begin{array}{cccc}
		\frac{-1}{\overline{\Omega}} & 0 & \frac{1}{\overline{\Omega}}&0\\
		0 & \frac{-1}{\overline{\Omega}} & 0 &\frac{1}{\overline{\Omega}} \\
	\end{array}\right]^T
\end{array}
\end{equation}
where $\overline{\Omega}$ is the maximum angular speed the wheels' motors can perform, and $\mathbf{1}$ denotes a vector of proper dimension containing all ones. 

The differential-drive kinematics (\ref{eq:diffdrive-discrete}) can be transformed into equivalent unicycle kinematics via the following change of input variables: 
\begin{equation}\label{eq:diff-uni-transformation}
\left[
\begin{array}{c}
v^i(k)\\
\omega^i(k)
\end{array}
\right]
=
T
\left[
\begin{array}{c}
\omega^i_R(k)\\
\omega^i_L(k)
\end{array}
\right],\quad T=\left[
\begin{array}{cc}
\frac{r}{2}&\,\,\,\,\frac{r}{2}\\
\frac{r}{D}&-\frac{r}{D}
\end{array}
\right]
\end{equation}
obtaining:
    \begin{equation}\label{eq:unicycle-disrete}
    	\begin{array}{rcl}
    		x^i(k+1)&=&x^i(k)+T_s v^i(k)\cos \theta^i(k)\\
    		y^i(k+1)&=&y^i(k)+T_s v^i(k)\sin \theta^i(k)\\
    		\theta^i(k+1)&=&\theta^i(k)+T_s\omega^i(k)
    	\end{array}
    \end{equation}
where 
$v^i,\omega^i\in\rr$ are the linear and angular speeds of the robot respectively. 

The input constraint set \eqref{eq:diff-drive-constraints}, mapped into the unicycle input space, transforms into a rhombus-like set, $\mathcal{U}_u\subset \rr^2,\,\,{0_2=[0,0]^T}\in \mathcal{U}_u,$ which defines the admissible linear and angular velocities for the unicycle, i.e.,
\begin{equation}\label{eq:input-constraint-unicycle}
\mathcal{U}_u\!=\!\{[v,\omega]^T\in\rr^2\!: H_u\left[v^i,\omega^i \right]^T\leq {\bf{1}}\},\, H_u=H_dT^{-1}
\end{equation}
%
% \todo[inline,color=cyan]{Fig. \ref{fig:input_constraints} needs to be edited:

%     1) add the letter to each subplot (a)-(b)-(c)
    
%     2) add the sets name
    
%     3) add the index $i$ to the variables
    
% }
% \begin{figure}[!h]
% 	\centering
% \includegraphics[width=1\linewidth]{Figure/input_constraints.eps}	\caption{Sets of feasible control inputs for differential-drive (a), unicycle (b), and feedback-linearized dynamics (c)}
% 	\label{fig:input_constraints}
% \end{figure}%
%

\subsection{Formation Setup and Problem Formulation}\label{sec:prob-form}

%Let's consider a formation of autonomous mobile robots. The platooning control problem considered in this paper consists of designing state-feedback control laws for each agent that drive the formation along a desired online-generated trajectory while keeping a platoon configuration. 

\noindent \textit{Considered setup}: Consider a formation of $N$ mobile robots (i.e., the agents) described by the constrained kinematic model \eqref{eq:diffdrive-discrete}-\eqref{eq:diff-drive-constraints}. The agents are organized in a leader-followers configuration, where $i=0$ denotes the index of the leader robot and $i=1\dots N-1$ the indexes of the followers. 

\begin{figure}[!h]
	\centering
\includegraphics[width=1\linewidth]{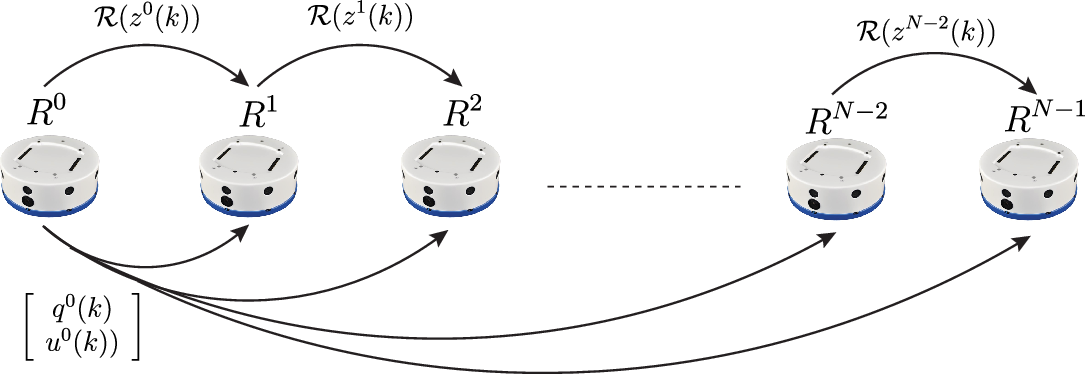}
	\caption{Vehicle to Vehicle (V2V) communication graph}
	\label{fig:V2V}
\end{figure}

We assume that the leader agent is equipped with an online path planner providing 
a  bounded and smooth 2D-trajectory in terms of reference position $(x_r(t),y_r(t))$, velocities $(\dot{x}_r(t),\dot{y}_r(t)$, and accelerations $(\ddot{x}_r(t),\ddot{y}_r(t))$ for the leader robot's geometric center, where $t\in \rr^+.$ Then the leader's pose and control inputs are broadcasted to all the follower's agents. To this end, different communication channels are established, i.e., between the leader agent and the followers, and between two consecutive agents $i$ and $i+1$, $\forall i=0,1,\dots N-2$ (see the network topology in Fig.\ref{fig:V2V}). The latter requirement is essential to guarantee collision avoidance capabilities between subsequent agents.
We also assume that the leader's path planner module is capable of generating a safe trajectory that does not intersect the followers' positions, with a certain safe distance $\overline{d}$, at any given time, i.e., 
    $$
        dist(z_r(k),\mathcal{B}(\overline{d},z^i(k))>0,\forall k\geq0,\forall i=1,\dots,N-1        
    $$
    where $$\mathcal{B}(\overline{d},z^i(k))=\{z^i \in \rr^2 \mid(z^i-z^i(k))^TQ_{\overline{d}}^{-1}(z^i-z^i(k)) \leq 1\}$$
    with $ Q_{\overline{d}} = \overline{d}^2 I$.
Moreover, the reference longitudinal velocity is assumed to be lower bounded by $\underline{v_r}$, i.e.,  $v_r(k)>\underline{v_r},\,\forall k\geq 0$
 \noindent All the vehicles are required to follow the same reference trajectory with a desired inter-vehicles delay $\overline{\eta}^i>0$ where $\overline{\eta}^i>\overline{\eta}^j$ if $\forall i>j$
\begin{remark}
   In order to guarantee collision-avoidance requirements, the inter-agent delay is assumed to be dynamically adjustable at runtime. To this end, in the following, the inter-agent delay is treated as a function of time $k$, namely $\eta^i(k)>0.$  
\end{remark}

\begin{problem}\label{problem}
\it   Given the reference pose $q_r(k)=\left[x_r(k),y_r(k),\theta_r(k)\right]^T$ and the setup described above, design a platooning control strategy such that all the agents can track a delayed reference trajectory while ensuring absence of collisions. Consequently, the leader and follower subproblems of interest are:

    \textit{\textit{[P1-1]}}:
     \textit{Design a trajectory tracking control law $[\omega^0_{R}(k),\omega^0_{L}(k)]^T=\phi^0(k,q^0(k),q_r(k))\in\mathcal{U}_d$ such that the tracking error of the leader with respect to the reference trajectory, namely $\tilde{q}^0(k)=q^0(k)-q_r(k)$ remains bounded $\forall k\geq 0$}.

    \textit{\textit{[P1-2]}}:
      {\it Design a trajectory tracking control law $[\omega^i_{R}(k),\omega^i_{L}(k)]^T=\phi^i(k,q^i(k),q^0(k-\eta^i(k)))\in\mathcal{U}_d$ such that $\tilde{q}^i(k)=q^i(k)-q^0(k-\eta^i(k))$ remains bounded $\forall k\geq 0, \, \forall i=1\dots N-1$, i.e. the followers track the leader pose delayed of $\eta^i(k)\in \mathbb{N}^+$ time instants while avoiding collisions with the other robots, where $\eta^i(k)$ is the \textit{inter-vehicle delay} for the agent $i.$}
\end{problem}

\begin{remark}
The robot's reference orientation $\theta_r(t)$, longitudinal velocity $v_r(t)$ and angular velocity $\omega_r(t)$, fulfilling the unicycle kinematics, can be computed as \cite{oriolo2002wmr}: 
\begin{equation}\label{eq:ref-traj-variables}
	\begin{array}{rcl}
	    \left[
	    \begin{array}{c}
	         v_r(t) \\
	         \omega_r(t)
	    \end{array}
	    \right]&=&
	    \left[
	    \begin{array}{c}
 \sqrt{\dot{x}_r(t)^2+\dot{y}_r(t)^2}\\
	    \frac{\ddot{y}_r(t)\dot{x}_r(t)-\ddot{x}_r(t)\dot{y}_r(t)}{\dot{x}_r(t)^2+\dot{y_r}(t)^2}
	    \end{array}
	    \right]
	    \\
\theta_r(t)&=&\text{ATAN}_2\left(\dot{y}_r(t),\dot{x}_r(t)\right)
	\end{array}
\end{equation}
\end{remark}

%
% Let's consider a leader agent, performing an arbitrarily smooth and bounded trajectory following a predefined set of constraints on the angular velocities of the wheels of the robot, in the cartesian space with a position $(x_l(t),y_l(t))$, velocity $(\dot{x}_l(t),\dot{y}_l(t)$, and acceleration $(\ddot{x}_l(t),\ddot{y}_l(t)),$ where $t\in \rr^+.$. It is assumed that the leader shares the vital information on the wheels position, velocity, acceleration and angular velocity. 

% The follower robot is required to follow the leader robot at a constant distance ($d_l$). Also, there is a safety distance ($d_{safe}$) that is enforced to ensure that the follower robot never violates it and it is illustrated in fig. \ref{fig:distance_to_platoon}.

% \begin{figure}[!h]
% 	\centering
% \includegraphics[width=0.62\linewidth]{Figure/saftey_constant_distance.eps}
% 	\caption{Distance constraints in the platooning}
% 	\label{fig:distance_to_platoon}
% \end{figure}
\section{Proposed Solution} \label{sec:propo-solution}

In this section, the platooning formation control problem is solved by combining input-output feedback linearizations and set-theoretic MPC arguments. 
Specifically, first feedback linearization is used to derive an equivalent linear model describing the unicycle kinematics~\eqref{eq:unicycle-disrete}. Then, such a linearized model is exploited to derive a collision-free control strategy that drives the platoon along a desired reference trajectory. 

\subsection{Linearized Vehicle Kinematics via Input-Output Linearization}

By introducing the following change of output coordinates:
\begin{equation}\label{eq:input-outout-transf}
   z^i(k)=\left[\begin{array}{cc}
         x^i(k)+b\cos\theta^i(k),\,
	   y^i(k)+b\sin\theta^i(k) 
    \end{array}\right]^T 
    %\in \rr^2
\end{equation}
with $b>0$
i.e., representing the position of a point $\mathbf{B}^i$ displaced with respect to the geometric center of the robot, and by using the following input transformation
\begin{equation*}\label{feedback-linearization}
	\left[\!\!
	\begin{array}{c}
		v^i(k)\\
		\omega^i(k)
	\end{array}
	\!\!\right]
	\!=\!
	T_{FL}(\theta^i)\!
	\left[\!
	\begin{array}{c}
		u^i_1(k)\\
		u^i_2(k)
	\end{array}
	\!\right],\,\,	
	T_{FL}(\theta^i)\!=\!\left[\!\!\!
	\begin{array}{cc}
		\cos\theta^i&\sin\theta^i\\
		\frac{-\sin\theta^i}{b}&\frac{\cos\theta^i}{b} \\
	\end{array}\!\!
	\right]
\end{equation*}
the unicycle model (\ref{eq:unicycle-disrete}) is recast into the following two-single integrator model,
\begin{subequations}\label{eq:input-output-lin-sys}
\begin{gather}
    z^i(k+1)=Az^i(k)+Bu^i(k),\,\, A=I_{2\times2},\,\, B=T_sI_{2\times2} \label{eq:FL-model}\\
    \theta^i(k+1)=\theta^i(k)+T_s\frac{-\sin\theta^i(k)u^i_1(k)+\cos\theta^i(k)u^i_2(k)}{b} \label{eq:internal-dynamics}
\end{gather}
\end{subequations}

where $u^i(k)=[u^i_1(k),\,u^i_2(k)]^T\in \rr^2$ are the control inputs of linearized robot's model, while \eqref{eq:internal-dynamics} defines a decoupled nonlinear internal dynamics.

\subsection{Agents' Error Dynamics}

In this section, the feedback-linearized tracking error dynamics for each agent are derived. First, let's define the reference pose for the $i$-th agent as
\begin{equation*}
    q^i_r(k)=
    \begin{cases}    \left[x_r(k),y_r(k),\theta_r(k)\right]^T,\, \text{ If } i=0\;\\\\
        \!\left[x^0(k\!-\!\eta^i(k)),y^0(k\!-\!\eta^i(k)),\theta^0(k\!-\!\eta^i(k))\right]^T\\ \quad\quad\quad\quad \text{ If } i=1\dots N-1
    \end{cases}
\end{equation*}
where, $q^i_r(k) = [x^i_r(k),y^i_r(k),\theta^i_r(k)]^T$ and its reference control inputs as
\begin{equation*}
    \left[\begin{array}{c}
    v^i_r(k)\\ \omega^i_r(k)
    \end{array}\right]=\begin{cases}
        \left[v_r(k),\omega_r(k)\right]^T,\, \text{ If } i=0\;\\\\
        \left[v^0(k-\eta^i(k)),\omega^0(k-\eta^i(k))\right]^T,\\ \quad\quad\quad\quad  \text{ If } i=1\dots N-1
    \end{cases}
\end{equation*}
i.e, the reference is defined as the generated reference trajectory for the agent $0$, and as the leader delayed reference for all the agents $i=1\dots N-1$.
\begin{remark}
    The reference pose and inputs are assumed to satisfy the unicycle kinematics \eqref{eq:unicycle-disrete}, $\forall i=0,1\dots N-1$
\end{remark}
Then, the feedback linearized tracking error is defined as $\tilde{z}^i(k)=z^i(k)-z_r^i(k)$ where %
\begin{equation}\label{eq:lin-ref-traj}
	z_r^i=
	\left[\begin{array}{c}
		x^i_r(k)+b\cos\theta^i_r(k),\,
		y_r^i(k)+b\sin\theta^i_r(k)
	\end{array}\right]^T
\end{equation}
As shown in \cite{tiriolo2023set}, the linearized tracking error dynamics can be computed as follows:
\begin{equation}\label{eq:error-dynamics}
	     \tilde{z}^i(k+1)=A\tilde{z}^i(k)+Bu^i(k)+d^i(k)
\end{equation}
where \begin{equation}\label{eq:u_r}
    d^i(k)=-Bu^i_r(k), \quad u^i_r(k)=T_{FL}^{-1}(\theta^i_r(k))\left[v^i_r(k),\omega^i_r(k)\right]^T
    \end{equation}
%

% Similarly, the tracking error of the $i-th$ follower with respect to the linearized leader's delayed position can be defined as $$\tilde{z}^i(k)=z^i(k)-z^0(k-T_s\Bar{k}^i),\quad \forall i=1\dots N-1$$. It is straightforward to show that the error dynamics of the $i-th$ follower vehicles present the same structure as the leader's one, i.e., 
% %
% \begin{equation}\label{eq:error-dynamics-follower}   \tilde{z}^i(k+1)=A\tilde{z}^i(k)+Bu^i(k)+d^i(k)\quad \forall i=1\dots N-1
% \end{equation}
% %
% where 
% %
% \begin{equation}\label{eq:u_r_follower}
%     \begin{array}{rcl}
%     d^i(k)\!\!\!\!&=&\!\!\!\!-Bu^0(k-T_s\Bar{k}^i)\\ u^0(k-T_s\Bar{k}^i)\!\!\!\!&=&\!\!\!\!T_{FL}^{-1}(\theta^0(k-T_s\Bar{k}^i))\!\!\left[\!\!\begin{array}{c}
%     v^0(k-T_s\Bar{k}^i)\\
%     \omega^0(k-T_s\Bar{k}^i)
%     \end{array}\!\!\right]
% \end{array}
% \end{equation}

%
\begin{remark}\label{remark:disturbance_ball_bound}
\it
Under the assumption of bounded reference, the disturbance $d^0(k)$ is also bounded. Furthermore, since  $d^i(k),\, \forall i=0,1\dots N-1$, depends on the leader's control inputs and orientation which are assumed bounded, $d^i(k)\in \mathcal{D}^i\subset \rr^2, \, \forall i=0,1\dots N-1$. Moreover, knowing the bound of $d^i(k)$, $\mathcal{D}^i$ can be over-approximated with a ball of radius $r_{d^i},$ i.e, $\mathcal{D}^i=\mathcal{E}(Q^i_{d}),\,\,Q_{d}^i={r^i_{d}}^{2}I_{2\times2}$
%
% \begin{equation*}\label{eq:worst-case-disturbance}
% 	\mathcal{D}^i=\mathcal{E}(Q^i_{d}),\,\,Q_{d}^i={r^i_{d}}^{2}I_{2\times2}
% \end{equation*}
% \begin{equation*}\label{eq:worst-case-disturbance}
% 	\mathcal{D}^i=\{d^i\in\rr^2:d^i{^T}{Q_d^i}^{-1} d^i\leq 1\},\,\,Q_{d}^i={r^i_{d}}^{2}I_{2\times2}
% \end{equation*}
%\hfill $\Box$
\end{remark}
% \todo[inline,color=yellow]{the above $r_d^i$, when I change to the format, its causing some errors, unsure. I feel it is not uniform wrt mentions across the paper}
\begin{lemma} \cite{wang2003full}\label{lem:full-state-track}
\textit{If a control law $u(\cdot)$ is such that \eqref{eq:error-dynamics} is stable, the point $\bf B^i$ tracks any bounded reference trajectory with a bounded internal dynamic. Consequently, also the tracking error $\tilde{q}(k)$ is bounded. 
\hfill $\Box$
}
\end{lemma}

% \begin{figure}[!h]
% 	\centering
% \includegraphics[width=0.8\linewidth]{Figure/platoon_multi_khep_eg.eps}
% 	\caption{Platoon of 3 vehicles with inter-vehicle delay}
% 	\label{fig:distance_to_platoon}
% \end{figure}%

\subsection{Input Constraints Characterization}
In \cite{Tiriolo_TCST} it has been proved that the set of admissible inputs for the model \eqref{eq:FL-model}, and consequently for the error dynamics \eqref{eq:error-dynamics}, is the following orientation-dependent polyhedral set 
\begin{equation}\label{eq:tv-input-constraints}
\begin{array}{cc}
    &\mathcal{U}(\theta^i)=\{\left[u^i_1,u^i_2\right]^T\in\rr^2:\, H(\theta^i)\left[u^i_1,u^i_2\right]^T\leq {\bf{1}}\},\\
    &H(\theta^i)=H_dT^{-1}T_{FL}(\theta^i)=\\
    &=\left[\begin{array}{cc}
        \frac{D\sin\theta^i - 2\cos\theta^i b}{2\overline{\Omega} R b} & \frac{-D\cos\theta^i - 2\sin\theta^i b}{2\overline{\Omega} R b} \\
        \frac{-D\sin\theta^i - 2\cos\theta^i b}{2\overline{\Omega} R b} & \frac{D\cos\theta^i - 2\sin\theta^i b}{2\overline{\Omega} R b}\\
        \frac{-D\sin\theta^i +2\cos\theta^i b}{2\overline{\Omega} R b} & \frac{D\cos\theta^i + 2\sin\theta^i b}{2\overline{\Omega} R b} \\
        \frac{D\sin\theta^i + 2\cos\theta^i b}{2\overline{\Omega} R b} & \frac{-D\cos\theta^i + 2\sin\theta^i b}{2\overline{\Omega} R b}
    \end{array}\right]
\end{array}
\end{equation}
It has also been proved that there exists a worst-case circular inner approximation $\mathcal{U}(\theta^i),\, \forall\theta^i$, defined as follows:
\begin{equation}\label{eq:worst-case-constraint-explicit}	\hat{\mathcal{U}}^i=\displaystyle\bigcap_{\forall\theta^i}\mathcal{U}^i(\theta^i)=\mathcal{E}(Q^i_{u}),\,\,Q_u^i=r_u^{i2}I_{2\times2}
\end{equation}
where $r_u^i=\frac{2\overline{\Omega} R b }{\sqrt{4b^2+D^2}}$. 
Similarly, it can be proved there $\mathcal{U}(\theta^i)$ admits the following circular outer approximation, $\forall\theta^i$:
\begin{equation}\label{eq:outer-approx-input-constraints}	\tilde{\mathcal{U}}=\displaystyle\bigcup_{\forall\theta^i}\mathcal{U}(\theta^i)=\mathcal{E}(\Tilde{Q}_u),\,\,\Tilde{Q}_u=\Tilde{r}_u^2I_{2\times2}
\end{equation}
 where $  \Tilde{r}_u=\max\{\bar{\Omega} R,\,2\bar{\Omega} R b  \backslash D\}$

\subsection{Set-Theoretic Receding Horizon Control for Trajectory Tracking}

To address the trajectory tracking requirements imposed by the considered platooning tracking control problem \ref{problem}, we exploit the set-theoretic RHC proposed in \cite{tiriolo2023set} to solve a trajectory tracking control problem for input-output linearized mobile robot described by \eqref{eq:error-dynamics}. Such a strategy can be summarized as follows.

\noindent
\textit{Notation: in the following,  $\mathcal{T}^i_j$ denotes the $j$-th set for the $i$-th robot.} 

{ The algorithm consists of two distinct phases:
\newline
\textit{- Offline:} For each agent $i=0,1,\dots N-1$, first, define the 
%design an 
optimal state-feedback control law $u^i(k)=-B^{-1}\Tilde{z}^i(k)$
%
% \begin{equation}\label{eq:terminal-control-law}
%     u^i(k)=-B^{-1}\Tilde{z}^i(k)
% \end{equation}
% %
which ensures that
%such that 
the disturbance-free model is asymptotically stable.
%, i.e. the matrix $A-BK$ contains all the eigenvalues in the unit disk. 
Then, under the assumption $\mathcal{D}^i\subset B\hat{\mathcal{U}}^i$, the smallest RCI set $\mathcal{T}^i_0$ (see definition \ref{def:RCI}) associated with the feedback control law and with the worst-case input constraint $\hat{\mathcal{U}}^i$ is given by $\mathcal{T}^i_0=\mathcal{\mathcal{D}}^i$. Finally, starting from $\mathcal{T}^i_0$, build a family of ROSC sets $\{\mathcal{T}^i_j\}_{j=1}^{N^i_s},\, N^i_s>0$, until a desired region of the state-space is covered. 
\newline
\textit{- Online ($\forall\,k$):} First, compute the set-membership index $$j^i(k):={\displaystyle\min_{j^i}}\, s.t.\, \{j^i\geq0: \, \Tilde{z}^i(k)\in\mathcal{T}^i_j\}$$
Then:
\begin{itemize}
    \item if $j^i(k)>0$ solve:
    \begin{equation}\label{eq:online-optimization}
		\begin{array}{ccc}
			u^i(k)=\displaystyle\argmin_{u^i} J({\Tilde{z}^i}(k),u^i) \, \, s.t. \\
			A\tilde{z}^i(k)+Bu^i-d^i(k)  \in (\mathcal{T}^i_{j(k)-1})\\
            u^i\in\mathcal{U}(\theta^i(k))
		\end{array}
	\end{equation}
        where $J^i(\Tilde{z}^i(k),u^i)$ is a convex cost function.
        \item else $u^i(k)=-B^{-1}\tilde{z}^i(k)+\hat{u}^i_r(k)$
        where $\hat{u}^i_r(k)$ is computed such that $u^i(k)$ complies with the current input constraints, i.e.,
        \begin{gather}            \hat{u}^i_r(k)=\displaystyle\arg\min_{\hat{u}_r} \|\hat{u}^i_r-u^i_r(k)\|_2^2 \quad s.t.\label{eq:RCI-optim-obj}\\
				-B^{-1}\tilde{z}^i(k)+\hat{u}^i_r\in \mathcal{U}(\theta^i(k))\label{eq:RCI-optim-constr}	
			\end{gather}
\end{itemize}
% \todo[inline,color=yellow]{I think $\hat{u}^i_r(k)$ is computed such that $\hat{u}^i(k)$ satisfies the input constraints, I am not sure how the above statement conveys that?}
%
\begin{property}
In \cite{tiriolo2023set}, it has been proved that
\begin{itemize}
    \item $\forall \tilde{z}^i(0)\in\bigcup_{j=0}^N\mathcal{T}^i_j\, \forall d^i(k)\in\mathcal{D}^i$, the tracking-error state trajectory is Uniformly Ultimately Bounded (UUB) in $\mathcal{T}^i_0$, i.e., there exists a sequence of at most $N^i_s$ control inputs that brings $\Tilde{z}^i(k)$ into the terminal set $\mathcal{T}^i_0$. 
    \item Optimization \eqref{eq:online-optimization} is recursively feasible by construction. 
    \item The offline computed terminal feedback control law , is optimal with respect to a Linear-Quadratic (LQ) cost. 
    \item Optimization \eqref{eq:RCI-optim-obj}-\eqref{eq:RCI-optim-constr} provides the best feed-forward term $u_r(k)$, compatible with the time-varying input constraints, that cancels out, completely or partially, the effect of the disturbance $d^i(k)$.
    \item Given the circular structure of the sets $\hat{\mathcal{U}}^i$ and $\mathcal{D}^i$,  starting from $\mathcal{T}^i_0$ a family of circular ROSC sets can be built as follows:
    \begin{equation}
        \!\! \!\!\!\mathcal{T}^i_j=\mathcal{E}(Q^i_{j}), \, Q^i_j={r_j^{i}}^2I_{2\times2},\,
    	r_j^i=r_{j-1}^i-r_d^i+T_s r_u^i \label{eq:one-step-T_i}
    \end{equation}
    % \begin{gather}
    % 	\mathcal{T}^i_j=\mathcal{E}(Q^i_{j}), \, Q^i_j={r_j^{i}}^2I_{2\times2} \label{eq:one-step-T_i}
    % 	\\
    % 	r_j^i=r_{j-1}^i-r_d^i+T_s r_u^i \label{eq:one-step-Q_i}
    % \end{gather}
    %
    and 
     the set-membership signal $j(k)$ can be computed as $j(k):=\min\{j:\Tilde{z}^i(k)^TQ_j^{i^{-1}}\Tilde{z}^i(k)\leq 1\}$
    %
    % \begin{equation}\label{eq:set-membership-cond}
    % 		j(k):=\min\{j:\Tilde{z}^i(k)^TQ_j^{i^{-1}}\Tilde{z}^i(k)\leq 1\}
    % \end{equation}
\end{itemize}
\end{property}
\begin{remark}
    Since the bound of the set $\mathcal{D}^i$ depends on the reference trajectory (see Eq. \eqref{eq:u_r}), the containment condition $B\mathcal{U}\subset\mathcal{D}^i,\,\forall$ imposes a constraint the linear and angular velocity of the reference trajectory. Moreover, to guarantee that the containment condition is satisfied for each follower robot, the input constraint set of the leader, namely $\hat{\mathcal{U}}^0$, must be such that $\hat{\mathcal{U}}^0\subset\hat{\mathcal{U}}$
\end{remark}

}

\subsection{Proposed Predictive Platooning Tracking Control}
In the following, the above-discussed predictive control strategy is extended to solve the platooning control problem \ref{problem} (see steps \ref{alg1-step-ST1}-\ref{alg1-step-ST2} of algorithm \ref{alg:leader} and steps \ref{alg2-step-ST1}-\ref{alg2-step-ST2} of algorithm \ref{alg:follower}). 
To this end, in the following, the leader's and follower's control algorithms are addressed separately. 
In order to provide formal guarantees of the absence of collisions between the agents a further assumption on the initial formation's configuration is needed. 
\begin{assumption}\label{ass:initial-conf}
    Initial spatial configurations are sequentially assigned to agents depending on their indexes, i.e., 
    \begin{equation}
        \begin{array}{c}
            \|z^0(0)-z^1(0)\|_2^2<\|z^0(0)-z^2(0)\|_2^2<\dots\\\dots <\|z^0(0)-z^{N-1}(0)\|_2^2
        \end{array}
    \end{equation}
\end{assumption}
%
% \begin{assumption}
%     The agents are equipped with vehicle-to-vehicle (V2V) communication modules allowing the leader robot to transmit its pose and control inputs to each follower, and allowing each robot to transmit its one-step reachable set to its immediate predecessor (refer to Fig. \ref{fig:V2V} for further details). 
% \end{assumption}

\noindent
\textit{\textit{1) Leader's control strategy:}}
%
% \begin{lemma}\label{lem:leader-problem}
%  Algorithm \ref{alg:leader} solves Problem P1-1.   
% \end{lemma}
% \begin{proof}
%     See \cite{tiriolo2023set}.
% \end{proof}
%
%
%
\begin{algorithm}[!h]
	\caption{Leader's Tracking Algorithm}\label{alg:leader}
	\vspace{0.1cm}
	\noindent	\textit{Offline:}
	\begin{algorithmic}[1]
		\State Set $\mathcal{U}=\hat{\mathcal{U}}^0$ with $\hat{\mathcal{U}}\subset \hat{\mathcal{U}}^0 \subset B^{-1}\mathcal{D}^0$, $K=B^{-1},$ and  $\mathcal{T}^0_0=\mathcal{D}^0;$ Build $\{\mathcal{T}^0_j\}_{j=1}^{N_s}$  using \eqref{eq:one-step-T_i}; Store $\{\mathcal{T}^0_j\}_{j=0}^{N_s}.$
	\end{algorithmic}
	\vspace{0.1cm}
	\textit{Online:}

 \begin{algorithmic}[1]
		\State  Measure $q^0(k)$ and compute $\Tilde{z}^0(k)=z^0(k)-z^0_r(k)$, with $z^0(k)$ as in  \eqref{eq:input-outout-transf}, $z^0_r(k)$ as in \eqref{eq:lin-ref-traj};\label{alg1-step-1}
        \State Send $q^0(k),v^0(k),\omega^0(k)$ to each agent $i=1,\dots,N-1$
		\State  Compute $\mathcal{U}(\theta^0)$ as in \eqref{eq:tv-input-constraints} and $u^0_r(k)$ as in \eqref{eq:u_r};
		\State Compute $j(k):=\min\{j:\Tilde{z}^0(k)^TQ_j^{0^{-1}}\Tilde{z}^0(k)\leq 1\}$
		\If{$j(k)>0,$} $u^0(k)=\displaystyle\arg\min_{u} J(x,u)\,\,s.t.$ \label{alg1-step-ST1}
		\begin{subequations}\label{eq:prop-online-optimiz-ROSC}
			\begin{gather}
				A\tilde{z}^0(k)+B{u}^0-Bu^0_r(k) \in \mathcal{T}^0_{i(k)-1}, \,\,
				u\in \hat{\mathcal{U}}^0 \nonumber
			\end{gather}
		\end{subequations}
		\Else{ $u^0(k)=-B^{-1}\tilde{z}^0(k)+\hat{u}^0_r(k), \quad \text{where}$} 
  % \begin{equation}\label{eq:terminal-law-leader}
		    
		% \end{equation} 
		%
        \vspace{-0.6cm}
		\begin{subequations}\label{eq:prop-online-optimiz-RCI-leader}
			\begin{gather}
\hat{u}^0_r(k)=\displaystyle\arg\min_{\hat{u}_r} \|\hat{u}_r-u^0_r(k)\|_2^2 \quad s.t.\nonumber\\
				-B^{-1}\tilde{z}^0(k)+\hat{u}_r\in \hat{\mathcal{U}}^0 \nonumber
			\end{gather}
		\end{subequations}
		\EndIf \label{alg1-step-ST2}
		\State  Compute $\left[\omega^0_R(k),\omega^0_L(k)\right]^T=T^{-1}T_{FL}(\theta^0(k)){u}^0(k)
		$
		and apply it to the robot;  $k\leftarrow k+1$, go to \ref{alg1-step-1};
	\end{algorithmic}
\end{algorithm}
\noindent
\textit{\textit{2) Follower's control strategy:}}
First, it is worth noting that Problem P1-1 is equivalent to Problem P1-2 where $q_r=q^0(k-\eta^i(k))$, i.e., the reference trajectory is replaced with the leader's 
pose delayed of $\eta^i(k)$ time instants. To this end, Algorithm \ref{alg:leader}, can be extended to solve Problem P1-2. However, although the online planner module (see Section \ref{sec:prob-form} ensures the absence of collisions between the leader robot and each follower, the possibility of collision between followers may arise. 
To provide collision-free guarantees, in the following, ROSC sets are exploited to ensure there are no intersections between the trajectories performed by the agents. 
%To this end, the following control policy is proposed for the follower robots in order to avoid collisions. 
 By denoting with $z^i(k)$ the $i$-th follower robot's position, and with $\mathcal{R}^i(z^i(k))$ its one step-reachable set starting from the point $z^i(k)$ (see Definition \ref{Def:ROSR}), a collision-avoidance policy can be stated as follows:
\begin{equation}\label{eq:collision-avoidance}
\begin{array}{c}
     \text{if }  \mathcal{R}^i(z^i(k))\bigcap \mathcal{R}^{i-1}(z^{i-1}(k))\neq \emptyset \implies \\ 
  u^i(k)=0,\, \eta^i(k)\leftarrow \eta^i(k)+1,\, \forall i=1,\dots N-1
\end{array} 
\end{equation}
i.e., if at any time $k\geq 0$, the ROSR of agent $i$ intersects the one of its immediate predecessor 
then the agent $i$ is stopped and its inter-agent delay $\eta^i(k)$ incremented. 
Such a policy ensures that the robots' trajectories never overlap. 
%
% \begin{figure}[!h]
% 	\centering
% \includegraphics[width=1\linewidth]{Figure/collision-avoidance-policy.eps}
% 	\caption{Collision-free (a) and potential collision (b) scenarios}
% 	\label{fig:collision-avoidance}
% \end{figure}%
%
%
% \begin{remark}
%     By applying Definition \ref{Def:ROSR}, the ROSR set from the point $z^i(k)$ is given by:
%     %
%     \begin{equation*}
%         \mathcal{R}(z^i(k)) = \{ z\in\rr^n : z \in A  z^i(k)\oplus (B\cdot \mathcal{U}(\theta^i(k))) \oplus \mathcal{D}^i\}
%     \end{equation*}
%     %
%     where $\mathcal{U}(\theta^i(k))$ is a time-varying polyhedral set. Therefore, the computation of $\mathcal{R}(z^i(k))$ requires a numerical procedure. 
%     In order to provide an analytical way to compute such ROSR sets, we over-approximate $\mathcal{U}(\theta^i(k))$ with a circular set $\Tilde{\mathcal{U}}$.
%     Then, the ROSR set $\mathcal{R}^i(z^i(k))$ is the following circular set:
%     %
%     \begin{equation}\label{eq:ROSR-computation}
%         \begin{array}{c}
%              \mathcal{R}^i(z^i(k))\!\!=\!\{z^i\!\in\!\rr^2\!:\!{ (z^i-z^i(k))^{T}}{Q^i_R}^{-1} 
%  \!(z^i-z^i(k))\!\leq \!1\!\}
% 	\\
% Q^i_R={r^i_R}^2I_{2\times2}, \quad 	r^i_R=r^i_d+T_s \Tilde{r}^i_u 
%         \end{array}
%     \end{equation}

% \end{remark}
% %
% \todo[inline, color=yellow]{The above equations require index term i, not sure if the definition is right, please verify. I think it should be defined as $\mathcal{R}^i(z^i(k))=\{z^i\in\rr^2:{ (z^i-z^i(k))^{T}}Q_R^{-1} 
 % (z^i-z^i(k))\leq 1\}$}
{
It can be proved that Algorithms \ref{alg:leader}-\ref{alg:follower} solve Problem \ref{problem}.

}
\begin{algorithm}[!h]
	\caption{$i-$th follower's Tracking Algorithm}\label{alg:follower}
	\vspace{0.1cm}
	\noindent	\textit{Offline:}
	\begin{algorithmic}[1]
		\State Set $\mathcal{U}=\hat{\mathcal{U}}$, $K=B^{-1},$ and  $\mathcal{T}^i_0=\mathcal{D}^i;$ Build $\{\mathcal{T}^i_j\}_{j=1}^{N_s}$  using \eqref{eq:one-step-T_i}; Store $\{\mathcal{T}^i_j\}_{j=0}^{N_s}.$
	\end{algorithmic}
	\vspace{0.1cm}
	\textit{Online:}
	\begin{algorithmic}[1]
		\State  Measure $x^i(k),\,y^i(k), \,\theta^i(k)$ and compute $\Tilde{z}^i(k)=z^i(k)-z^i_r(k)$, with $z^i(k)$ as in  \eqref{eq:input-outout-transf}, $z^i_r(k)$ as in \eqref{eq:lin-ref-traj};\label{alg2-step-1}
        \State Receive $q^0(k),\, v^0(k),\,\omega^0(k)$ from the leader and store them into a sequence;
        \State Compute $\mathcal{R}^i(z^i(k))$ and send it to agent $i+1$
        \State Receive $\mathcal{R}^{i-1}(z^{i-1}(k))$ from agent $i-1$;
        \If{$\mathcal{R}^i(z^i(k))\bigcap \mathcal{R}^{i-1}(z^{i-1}(k))\neq\emptyset$}
        $$
        u^i(k)=0,\quad \eta^i(k)\leftarrow \eta^i(k)+1
        $$
        \Else{
          Compute $\mathcal{U}(\theta^i)$ as in \eqref{eq:tv-input-constraints} and $u_0(k)$ as in \eqref{eq:u_r};
		\State Compute $j(k):=\min\{j:\Tilde{z}^i(k)^TQ_j^{i^{-1}}\Tilde{z}^i(k)\leq 1\}$;
            \If{$j(k)>0,$} \label{alg2-step-ST1}
	\begin{subequations}\label{eq:prop-online-optimiz-ROSC-follower}
			\begin{gather}
				u^i(k)=\displaystyle\arg\min_{u} J(\Tilde{z}^i(k),u^i)\,\,s.t.\nonumber\\
				A\tilde{z}^i(k)+B{u}^i-Bu^0(k-\eta^i(k)) \in \mathcal{T}^i_{i(k)-1}, \,\,
				u\in \hat{\mathcal{U}}^i\nonumber
			\end{gather}
		\end{subequations}
		 \Else{ $u^i(k)=-B^{-1}\tilde{z}^i(k)+\hat{u}^i_r(k), \quad \text{where}$} 
   %\begin{equation}\label{eq:terminal-law-follower}
		    
		% \end{equation} 
		%
        \vspace{-0.35cm}
		\begin{subequations}\label{eq:prop-online-optimiz-RCI-follower}
			\begin{gather}
		\hat{u}^i_r(k)=\displaystyle\arg\min_{\hat{u}_r} \|\hat{u}_r-u^0(k-\eta^i(k))\|_2^2 \quad s.t. \nonumber\\
				-B^{-1}\tilde{z}^i(k)+\hat{u}_r\in \hat{\mathcal{U}}^i\nonumber
			\end{gather}
		\end{subequations}
		\EndIf\label{alg2-step-ST2}
        
        }
        \EndIf
		\State  Compute $
			\left[\omega^i_R(k),\omega^i_L(k)\right]^T=T^{-1}T_{FL}(\theta^i(k)){u}^i(k)$
		and apply it to the robot;  $k\leftarrow k+1$, go to \ref{alg2-step-1};
	\end{algorithmic}
\end{algorithm}

\section{Experimental Results} \label{sec:Exp_results}
% \begin{figure}[!h]
%  	\centering
%         \includegraphics[width=0.8\linewidth]{Figure/test_bed.jpg}
%  	\caption{Test-bed consisting of three Khepera robots and Vicon localization}
%  	\label{fig:test_bed}
%  \end{figure}
\begin{figure*}[!h]
  \centering
\includegraphics[width=1\textwidth]{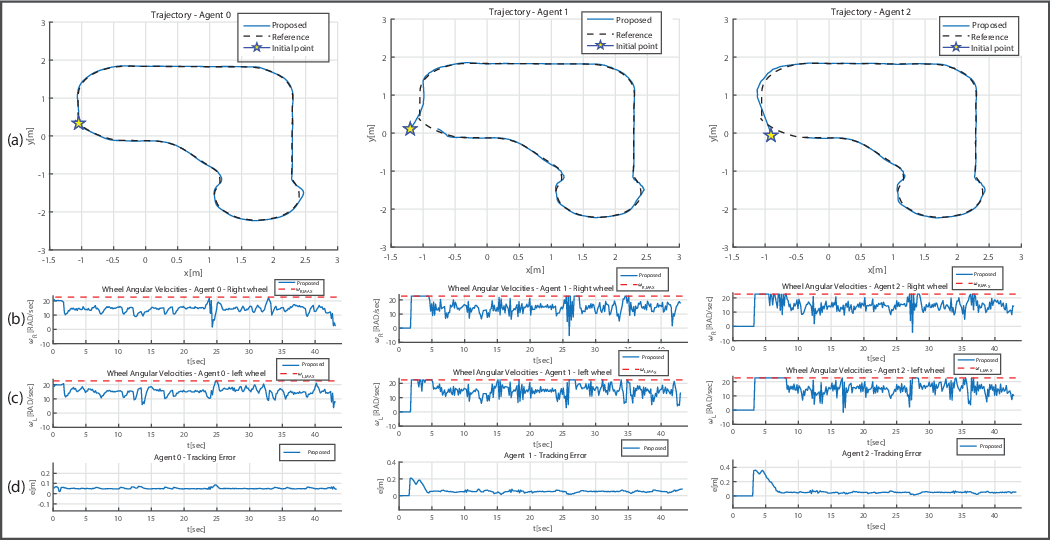}
  \caption{Experimental results: performed trajectory, wheel's angular velocity, and tracking error for agents $i=0,1,2$.}
  \label{fig:results}
\end{figure*}
The proposed platooning control strategy has been  validated through hardware-in-the-loop experiments, conducted with a platoon of $N=3$ Khepera IV differential drive robots. 
A demo of the proposed experiment can be found at the following web link: \url{https://youtu.be/UFS2VQJUQQo?si=5xZCv0hLS15ut44f}.

Each robot consists of two independently-driven wheels of radius $R =0.021[m]$, and axis length $D=0.1047 [m]$, capable of performing a maximum angular velocity $\overline{\Omega}=1200 [steps/sec]=38.71[rad/sec]$. However, to avoid unmodeled dynamic effects the maximum allowed angular velocity has been reduced to $\overline{\Omega}=700 [steps/s]=22.5833[rad/sec]$. Furthermore, the robots' kinematics have been feedback-linearized using $b=0.1$ and discretized using a sampling time $T_s=0.15$.
An ad-hoc Indoor Positioning System (IPS) has been realized using a Vicon motion capture system and an unscented Kalman Filter algorithm \cite{wan2001unscented}, which is capable of providing accurate measurements of each agent's pose. The control strategy has been implemented on a workstation equipped with an \textit{Intel 17-12700F} processor, running  Matlab 2022b. Each robot communicates with its own controller through a TCP communication channel.

\noindent
\textit{Control algorithm configuration:}
The proposed strategy has been configured with the following parameters: $J(\tilde{z},u)=\|A\tilde{z}(k)+B{u}\|_2^2$, $r_{d}^0 = 0.0507 [m]$, $r_u^0 = 0.40 [m]$,  $r_{d}^i = 0.40[m]$, $r_u^i = 0.4202[m]$, $\forall i=1\dots N-1$, $\tilde{r}_u = 0.9059 [m]$, $N_s^i=1000$.
% An admissible reference trajectory, complying with the assumptions made in Sec.\ref{sec:prob-form} is generated by a planner module running on the same workstation. The generated reference trajectory spans across an area of $3.5[m] \times 4.25[m]$, and it is characterized by a constant velocity along the path (approximately $0.32 [m/sec])$. 
An admissible reference trajectory, complying with the assumptions made in Sec.\ref{sec:prob-form} is generated by a planner module by interpolating a set of waypoints distributed along a path using cubic spline with the average longitudinal velocity between two subsequent waypoints equal to $0.32 [m/sec]$ and the generated reference trajectory spans across an area of $3.5[m] \times 4.25[m]$. In order to evaluate the tracking performance, the integral absolute error (IAE) ($\int_{0}^{k_f}|e(k)|dk$) is used, where $e(k) =\sqrt{(x^i_r(k)-x^i(k))^2+(y_r^i(k)-y^i(k))^2}$.
The results of the performed experiment are shown in Fig.~{\ref{fig:results}. Specifically, Fig.~ \ref{fig:results} shows the trajectory performed by the agents (a), angular velocities generated by the control algorithms (b)-(c), and considered tracking error $e(k)$ (d). Specifically, it can be appreciated how the proposed control algorithm is capable of ensuring a small tracking error for all the agents.   Furthermore, in Figs.~\ref{fig:results}(b)-(c), it can be appreciated how the control inputs fulfill the prescribed input constraints. Moreover, the following values of IAE  have been computed,  $2.2551$ , $2.2785$ and $2.7161$ for agents $0,1,2$ respectively. 
The obtained results confirm how the formation converges to a stable platoon configuration, i.e., the inter-agents delay $\eta(k)^i\rightarrow\overline{\eta}^i$, as $k\rightarrow\infty, \forall i=0,1,2$.
}

% \begin{table}[!h]
% \caption{Tracking Performance index of Platoon}
% \resizebox{\columnwidth}{!}{%
% \begin{tabular}{|c|cc|cc|cc|}
% \hline
% \multirow{2}{*}{Index} & \multicolumn{2}{c|}{Agent-0} & \multicolumn{2}{c|}{Agent-1} & \multicolumn{2}{c|}{Agent-2} \\ \cline{2-7} 
%  & \multicolumn{1}{c|}{Proposed} & \cite{KLANCAR2011} & \multicolumn{1}{c|}{Proposed} & \cite{KLANCAR2011} & \multicolumn{1}{c|}{Proposed} & \cite{KLANCAR2011} \\ \hline
% IAE & \multicolumn{1}{c|}{2.2551} & 3.225 & \multicolumn{1}{c|}{2.2785} & 2.7738 & \multicolumn{1}{c|}{2.7161} & 3.3353 \\ \hline
% ISE & \multicolumn{1}{c|}{0.1206} & 0.2664 & \multicolumn{1}{c|}{0.1634} & 0.2802 & \multicolumn{1}{c|}{0.3708} & 0.5713 \\ \hline
% ITSE & \multicolumn{1}{c|}{2.6008} & 5.9389 & \multicolumn{1}{c|}{2.5286} & 3.7351 & \multicolumn{1}{c|}{3.5748} & 5.4430 \\ \hline
% ITAE & \multicolumn{1}{c|}{48.9147} & 72.0839 & \multicolumn{1}{c|}{46.3983} & 52.8883 & \multicolumn{1}{c|}{48.6321} & 55.7160 \\ \hline
% \end{tabular}
% }
% \label{tab:results}
% \end{table}

\section{Conclusions}\label{sec:conclusions}
In this paper, a novel control strategy has been proposed to address a platooning formation control problem for mobile robots. The proposed solution has been derived by combining feedback linearization and set-theoretic MPC arguments to achieve bounded trajectory tracking error for the considered platoon and deal with the input constraints of the considered mobile robots. Based on the concept of one-step forward reachable sets, a collision avoidance policy has been designed to guarantee the absence of collisions among agents. Finally, the proposed solution has been experimentally validated using a formation of Khepera IV robots. The obtained results show that the proposed solution achieves 
high performance in terms of formation tracking error. 
%and demonstrate the real-time applicability of the proposed control strategy. 
%
%%
% We presented a novel platooning control solution tailored for mobile robots. Our approach leverages input-output feedback linearization to transform the nonlinear dynamics of differential drive robots into a linear model. Notably, the strategy we proposed for both the leader and the follower takes into account time-varying input constraints, ensuring that the robotic agents track their desired trajectories.
% Furthermore, we introduced an ad-hoc collision avoidance algorithm, which not only guarantees collision-free operation but also contributes to maintaining the platoon in a stable configuration. The proposed collision avoidance algorithm uses the information on system evolution to provide a conservative strategy to mitigate a collision.

% To validate the effectiveness of our proposed algorithm, we conducted extensive experimental trials on a dedicated test-bed consisting of multiple Khepera IV differential drive robots. Additionally, we bench-marked our approach against an alternative platooning scheme. The results unequivocally demonstrate that our proposed algorithm outperforms existing methods, signifying its significant contributions to the field of mobile robotics and platooning control.
\bibliographystyle{IEEEtran}
\bibliography{references}        

% Generated by IEEEtran.bst, version: 1.14 (2015/08/26)
\begin{thebibliography}{10}
\providecommand{\url}[1]{#1}
\csname url@samestyle\endcsname
\providecommand{\newblock}{\relax}
\providecommand{\bibinfo}[2]{#2}
\providecommand{\BIBentrySTDinterwordspacing}{\spaceskip=0pt\relax}
\providecommand{\BIBentryALTinterwordstretchfactor}{4}
\providecommand{\BIBentryALTinterwordspacing}{\spaceskip=\fontdimen2\font plus
\BIBentryALTinterwordstretchfactor\fontdimen3\font minus
  \fontdimen4\font\relax}
\providecommand{\BIBforeignlanguage}[2]{{%
\expandafter\ifx\csname l@#1\endcsname\relax
\typeout{** WARNING: IEEEtran.bst: No hyphenation pattern has been}%
\typeout{** loaded for the language `#1'. Using the pattern for}%
\typeout{** the default language instead.}%
\else
\language=\csname l@#1\endcsname
\fi
#2}}
\providecommand{\BIBdecl}{\relax}
\BIBdecl

\bibitem{platoon_book}
A.~Sciarretta, A.~Vahidi \emph{et~al.}, \emph{Energy-efficient driving of road
  vehicles}.\hskip 1em plus 0.5em minus 0.4em\relax Springer, 2020.

\bibitem{Christian_review_platoon}
V.~Lesch, M.~Breitbach, M.~Segata, C.~Becker, S.~Kounev, and C.~Krupitzer, ``An
  overview on approaches for coordination of platoons,'' \emph{IEEE
  Transactions on Intelligent Transportation Systems}, vol.~23, no.~8, pp.
  10\,049--10\,065, 2022.

\bibitem{liu2022cooperative}
W.-J. Liu, H.-F. Ding, M.-F. Ge, and X.-Y. Yao, ``Cooperative control for
  platoon generation of vehicle-to-vehicle networks: a hierarchical nonlinear
  mpc algorithm,'' \emph{Nonlinear Dynamics}, vol. 108, no.~4, pp. 3561--3578,
  2022.

\bibitem{Subramani_mpc}
F.~Eiras, M.~Hawasly, S.~V. Albrecht, and S.~Ramamoorthy, ``A two-stage
  optimization-based motion planner for safe urban driving,'' \emph{IEEE
  Transactions on Robotics}, vol.~38, no.~2, pp. 822--834, 2022.

\bibitem{Franze_multi}
G.~Franzè, G.~Fedele, A.~Bono, and L.~D’Alfonso, ``Reference tracking for
  multiagent systems using model predictive control,'' \emph{IEEE Transactions
  on Control Systems Technology}, vol.~31, no.~4, pp. 1884--1891, 2023.

\bibitem{angeli2008ellipsoidal}
D.~Angeli, A.~Casavola, G.~Franz{\`e}, and E.~Mosca, ``An ellipsoidal off-line
  mpc scheme for uncertain polytopic discrete-time systems,''
  \emph{Automatica}, vol.~44, no.~12, pp. 3113--3119, 2008.

\bibitem{oriolo2002wmr}
G.~Oriolo, A.~De~Luca, and M.~Vendittelli, ``Wmr control via dynamic feedback
  linearization: design, implementation, and experimental validation,''
  \emph{IEEE Transactions on Control Systems Technology}, vol.~10, no.~6, pp.
  835--852, 2002.

\bibitem{Tiriolo_TCST}
C.~Tiriolo, G.~Franzè, and W.~Lucia, ``A receding horizon trajectory tracking
  strategy for input-constrained differential-drive robots via feedback
  linearization,'' \emph{IEEE Transactions on Control Systems Technology},
  vol.~31, no.~3, pp. 1460--1467, 2023.

\bibitem{tiriolo2023set}
C.~Tiriolo and W.~Lucia, ``A set-theoretic control approach to the trajectory
  tracking problem for input-output linearized wheeled mobile robots,''
  \emph{IEEE Control Systems Letters}, vol.~7, pp. 2347--2352, 2023.

\bibitem{borrelli_bemporad_morari_2017}
F.~Borrelli, A.~Bemporad, and M.~Morari, \emph{Predictive Control for Linear
  and Hybrid Systems}.\hskip 1em plus 0.5em minus 0.4em\relax Cambridge
  University Press, 2017.

\bibitem{wang2003full}
D.~Wang and G.~Xu, ``Full-state tracking and internal dynamics of nonholonomic
  wheeled mobile robots,'' \emph{IEEE/ASME Transactions on mechatronics},
  vol.~8, no.~2, pp. 203--214, 2003.

\bibitem{wan2001unscented}
E.~A. Wan and R.~Van Der~Merwe, ``The unscented kalman filter,'' \emph{Kalman
  filtering and neural networks}, pp. 221--280, 2001.

\end{thebibliography}
\end{document}